\documentclass{IEEEtran}

\usepackage{amssymb,amsfonts,amsmath}
\usepackage{ifthen}

\usepackage{multirow}
\newcommand{\comment}[1]{\textcolor{red}{#1}}
\usepackage{rotating}
\usepackage{tabu}
\usepackage{tikz}
\usepackage{authblk}

\newcommand{\refTable}[1]{Table~\ref{#1}} 
\newcommand{\refFigure}[1]{Figure~\ref{#1}}

\newcommand{\refEq}[1]{\textbf{eq}[\ref{#1}]}

\newcommand{\ie}[0]{\textit{i.e.},\ }
\newcommand{\eg}[0]{\textit{e.g.},\ }
\newcommand{\aka}[0]{\textit{a.k.a}.\ }

\newcommand{\uM}[0]{$\mu$M}

\newcommand{\uL}[0]{$\mu$L}

\newcommand{\hone}[0]{${}^1$\textsc{h}}
\newcommand{\bayesil}[0]{\textsc{bayesil}}
\newcommand{\BAYESIL}[0]{\textsc{bayesil}}
\newcommand{\BATMAN}[0]{\textsc{batman}}
\newcommand{\batman}[0]{\textsc{batman}}

\newcommand{\NMR}[0]{\textsc{nmr}}
\newcommand{\ppm}[0]{\textsc{ppm}}
\newcommand{\PPM}[0]{\textsc{ppm}}
\newcommand{\xh}[2]{
\ifthenelse{\equal{#1}{}}{x}{x[#1]}
\ifthenelse{\equal{#2}{}}{}{^{(#2)}}
}
\newcommand{\xhs}[2]{
\ifthenelse{\equal{#1}{}}{\mathbf{x}}{\mathbf{x}[#1]}
\ifthenelse{\equal{#2}{}}{}{^{(#2)}}
}
\newcommand{\cluster}[0]{{\scriptstyle {\mathcal{C}}}}
\newcommand{\clusters}[0]{{\mathbf{\scriptstyle {\mathcal{C}}}}}
\newcommand{\compound}[0]{{\scriptstyle {\mathcal{M}}}}
\newcommand{\compounds}[0]{{\mathbf{\scriptstyle {\mathcal{M}}}}}

\newcommand{\peak}[0]{\ensuremath{\mathsf{q}}}
\newcommand{\spec}[0]{\ensuremath{\mathsf{s}}}

\newcommand{\recspec}[0]{\ensuremath{\widehat{\mathsf{s}}}}
\def\predSpec#1#2{\recspec{(#2\,;#1)}}
\newcommand{\shift}[0]{\ensuremath{\delta}}
\newcommand{\shifts}[0]{\ensuremath{\boldsymbol{\delta}}}
\newcommand{\conc}[0]{\ensuremath{\rho}}
\newcommand{\concs}[0]{\ensuremath{\boldsymbol{\rho}}}
\newcommand{\var}[0]{\ensuremath{\mu}}
\newcommand{\vars}[0]{\ensuremath{\boldsymbol{\mu}}}

\newcommand{\y}[0]{\ensuremath{x}}
\newcommand{\Y}[0]{\ensuremath{{\mathcal{X}}}}
\newcommand{\pparam}[0]{\ensuremath{\theta}}
\newcommand{\pparams}[0]{\ensuremath{\boldsymbol{\theta}}}
\newcommand{\loss}[0]{\ensuremath{\mathsf{\ell}}}
\newcommand{\ff}{\ensuremath{\mathsf{f}}}
\newcommand{\kk}{\ensuremath{\mathsf{k}}}
\newcommand{\ttt}[1]{\ensuremath{^{(#1)}}}
\newcommand{\nn}[1]{\ensuremath{[#1]}}

\newcommand{\expp}[1]{ \exp \bigg \{ \; #1 \; \bigg \} }

\newcommand{\pp}{\ensuremath{\mathbb{P}}}

\newcommand{\I}[0]{\ensuremath{I}}

\newcommand{\w}{\omega}

\def\factorOf#1{\partial#1}

\long\def\comment#1{}

\def\rg#1{\textcolor{red}{ [[ RG says:\ #1\ ]]}}
\def\sr#1{\textcolor{blue}{ [[Siamak says:\ #1]]}}

\begin{document}



\title{Accurate, fully-automated NMR spectral profiling for metabolomics}




\renewcommand\Affilfont{\itshape\small}
\author[1,2]{Siamak Ravanbakhsh}
\author[1,3]{Philip Liu}
\author[1,3]{Trent C. Bjorndahl}
\author[1,3]{Rupasri Mandal}
\author[1]{Jason R. Grant}
\author[1]{Michael Wilson}
\author[1]{Roman Eisner}
\author[3]{Igor Sinelnikov}
\author[4]{Xiaoyu Hu}
\author[5]{Claudio Luchinat}
\author[1,2]{Russell Greiner}
\author[1,3,6]{David S. Wishart}
\affil[1]{Department of Computing Science, University of Alberta, Edmonton, AB, Canada}\affil[2]{Alberta Innovates Center for Machine Learning (AICML)}\affil[3]{Department of Biological Sciences, University of Alberta, Edmonton, AB, Canada}\affil[4]{Fiorgen Foundation, 50019 Sesto Fiorentino, Italy}\affil[5]{Centro Risonanze Magnetiche (CERM/CIRMMP),
University of Florence, 50019 Sesto Fiorentino, Italy}\affil[6]{National Research Council, National Institute for Nanotechnology (NINT)}


\maketitle


\begin{abstract}
  Many diseases cause significant changes to the concentrations of small molecules (\aka metabolites) that appear in a person's biofluids,
which means such diseases can often be readily detected from a person’s “metabolic profile”.
This information can be extracted from a person's biofluids using  \NMR\ spectroscopy.
Today, this is often done manually by trained experts, which means this process is relatively slow, expensive and can be error-prone.
A system that can quickly, accurately and autonomously produce a person's metabolic profile would enable efficient and reliable prediction of many such diseases from a single sample,
which could significantly improve the way medicine is practiced.

This paper presents such a system:
Given a 1D  $^1$H \NMR\ spectrum of a complex biofluid such as serum or CSF,
our \bayesil\ system can automatically determine this metabolic profile, and do so without any human guidance.
This requires first performing all of the required  spectral processing steps 
(\ie Fourier transformation, phasing, solvent-removal, chemical shift referencing, baseline correction, lineshape convolution)
 then matching this resulting spectrum against a reference compound library, which contains the “signatures” of each relevant metabolite.
Many of these processing steps are novel algorithms, and 
our matching step views spectral matching as an inference problem within a probabilistic graphical model that rapidly approximates the most probable metabolic profile.

Our extensive studies on a diverse set of complex mixtures
(real biological samples, defined mixtures and realistic computer generated spectra; each involving $\sim 50$ compounds),
show that \bayesil\  can autonomously and accurately find \NMR-detectable metabolites at concentrations as low as $2\mu$M,
in terms of both identification ($\sim 90\%$ correct) and quantification ($\sim 10\%$ error),
in under 5 minutes on a single CPU processor.
These results  demonstrate that \bayesil\  is the first fully-automatic publicly-accessible system that provides quantitative \NMR\ spectral profiling effectively -- 
with an accuracy that meets or exceeds the performance of highly trained human experts. 
We anticipate this tool will usher in high-throughput metabolomics and enable a wealth of new applications of NMR in clinical settings.
Users can access \bayesil\ at \textit{http://www.bayesil.ca}. 

\comment{  
In NMR-based metabolomics, spectral profiling or spectral
    deconvolution is frequently used to identify and quantify
    metabolites from 1D \hone NMR spectra. The process involves
    fitting the individual reference NMR spectra of many pure
    compounds to the NMR spectrum of a biofluid mixture.   
    Ideally what is needed is a system that extracts the identity and quantity of compounds
from the raw NMR spectra.
This system should 
automatically
    performs both spectral processing (\ie Fourier transformation,
    phasing, solvent filtering, referencing, baseline correction,
    reference lineshape convolution) and spectral profiling against
     a reference compound library.
    Furthermore, it should be able to analyze complex
    mixtures ($>50$ compounds) accurately ($>90\%$ correct) and yield
    compound concentration data that is $90\%$ accurate.  
    This report describes such a system, called
    \bayesil, which first uses a variety of novel phasing and baseline
    correction methods to automatically process 1D NMR spectra.  
    It then treats the spectral deconvolution as an inference problem
    within a probabilistic graphical model to rapidly approximate the most
    probable metabolic profile.
    Our extensive testing on defined mixtures, 
    real biological samples and computer generated spectra
    demonstrate that \bayesil\ operates at
    a level that meets or exceeds the performance of highly-trained
    human experts.  \bayesil\ appears to be the first system that
    supports fully automated and fully quantitative NMR-based
    metabolomics.    
    We anticipate this will open the door to high-throughput 
    metabolomics and routine
    applications of NMR in clinical settings.
 }
   \comment{ 
    Many diseases cause significant changes to the concentrations of small molecules (\aka metabolites) that appear in a person's biofluids, which means such diseases can often be readily detected from a person's ``metabolic profile''  -- 
    \ie the list of concentrations of those metabolites.  A tool that can quickly, accurately and autonomously produce a person's metabolomics profile would enable efficient and reliable prediction of many such diseases from a single sample, which could therefore revolutionize the way medicine is practiced.
This paper presents such a system:
Given a 1D $^1$H-NMR spectrum of a complex biofluid such as serum or cerebrospinal fluid (CSF),
our system, \bayesil, can automatically determine this metabolic profile, and do so without any human guidance.
Our extensive studies with defined mixtures,  
 real biological samples and realistic computer-generated spectra (each involving $\sim 50$ compounds)
show that \bayesil\ can autonomously both identify and quantify
metabolites with $\sim 90\%$ accuracy in under 5 minutes.
This requires first performing many spectral processing steps (\ie Fourier transformation, phasing, solvent filtering, referencing, baseline correction,  reference lineshape convolution), then matching this resulting spectrum against a reference compound library, which contains the ``signatures'' of each relevant metabolite.
Many of these processing steps required novel innovations to work autonomously, and our matching step views spectral deconvolution as an inference problem within a probabilistic graphical model to rapidly approximate the most probable metabolic profile.
These results  demonstrate that \bayesil\  is the first fully-automatic system that provides quantitative NMR spectral profiling effectively -- with an accuracy that meets or exceeds the performance of highly trained human experts. 
Users can access \bayesil\ at 
\textit{http://www.bayesil.ca}. 
 We anticipate this system will usher in high-throughput metabolomics and enable a wealth of new applications of NMR in clinical settings.
 }
 \comment{
}
 
 \end{abstract}






\section*{}
Metabolomics is a relatively new branch of ``omics'' science that
focuses 
on the system-wide characterization of 
metabolites~\cite{hmdb}.
\comment{
It is very relevant to medicine and healthcare as such small molecules are central to all life processes, 
providing the energy, the regulatory signals and the chemical building blocks that allow cells to grow, divide and interact with their environment. }%
Metabolomics is often viewed as 
complementary to the other 
``omics'' fields 
as it provides information about 
both an organism's phenotype and its environment.  
Because metabolomics provides a unique window on gene-environment interactions, it is playing an increasingly important role in many quantitative phenotyping and functional genomics studies~\cite{metabolomics_genetics1,metabolomics_genetics2,metabolomics_genetics3}.
It is also finding more 
applications in disease diagnosis, biomarker discovery and drug development/discovery (\eg \cite{metabolomics_biomarker2,metabolomics_biomarker4,metabolomics_biomarker5}).
This rapid growth in interest and excitement surrounding metabolomics is also
revealing its ``Achilles heel'':
Unlike proteomics, genomics or transcriptomics, which are \emph{high-throughput} sciences, 
metabolomics is a relatively 
\emph{low-throughput} science. 
Compared to genomics,
where it is now possible to automatically characterize 1000s of genes, 100’s of thousands of transcripts and millions of \textsc{snp}s in mere minutes, 
metabolomics only allows 
users 
to identify and measure a few dozen metabolites after many hours of manual effort.
In other words, 
\emph{metabolomics is not automated}. 
This problem may stem from the history of metabolomics,
as its analytical techniques,
\comment{
This lack of automation 
is probably because 
the analytical techniques used in metabolomics,
}%
such as \NMR\ spectroscopy, gas-chromatography-mass spectrometry~(\textsc{gc-ms}) 
and liquid chromatography-mass spectrometry~(\textsc{lc-ms}), were originally developed for identifying and quantifying 
{\em pure}\ compounds, not complex mixtures. 
Because most biological samples contain hundreds of metabolites, the resulting \NMR, \textsc{hplc} or \textsc{lc-ms} spectra usually contain hundreds or even thousands of peaks. The challenge in metabolomics, therefore, 
is to identify the mixture of compounds that produced this forest of peaks.  
This compound identification process,
called \emph{spectral profiling},
involves fitting the mixture spectrum to a set of individual pure reference spectra obtained from known compounds~\cite{qnmr,targetedprofiling}.
If done correctly,
the fitting process 
yields not only the identity of the compounds, but also 
the concentration of those compounds. Therefore, the end result of a successful spectral profiling study is a table of metabolite names and their absolute or relative 
concentrations. 
Because spectral profiling is such a complex pattern recognition problem, it is often best done by a trained expert.  
However, this reliance on 
manual data analysis by a human expert 
is
problematic, 
as it is slow and leads to inconsistent results, 
operator errors and reduced levels of reproducibility~\cite{between_person}.

\begin{figure*}
\includegraphics[width=1.0\linewidth]{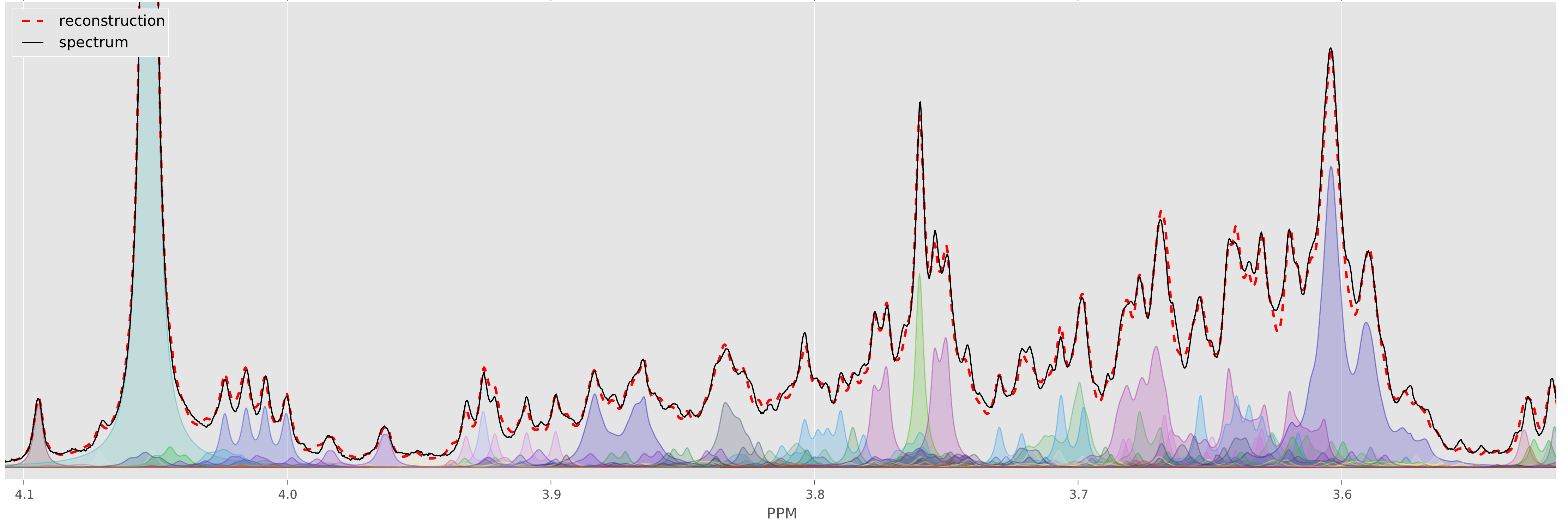}
\caption{This figure shows the crowded region (3.5-4.1 \PPM) of a computer generated spectrum with 150 compounds (solid black) 
and the fit produced by \bayesil\ (dashed red) as well as individual clusters as quantified by \bayesil.
Each cluster is free to shift a specified amount, which is at least 0.025 \PPM. 
}
\label{fig:fit}
\end{figure*}

The automation bottleneck in metabolomics is widely recognized,
and 
has led to 
a number of efforts to accelerate or automate compound identification and/or quantification in \textsc{lc-ms}, in \textsc{gc-ms} and in \NMR\ spectroscopy. 
Some of the most active efforts in (semi)automated compound identification and quantification have been in \NMR-based metabolomics. 
In particular, several software packages have been developed that support semi-automatic \NMR\ spectral profiling of 1D and 2D 
\hone~\NMR\ spectra, 
including some commercial packages 
(\eg \cite{autofit,ceed,batman}).
There are tremendous differences in the quantification/identification capabilities, spectral database size, speed and instrument compatibility of these software tools.
Furthermore, 
 these packages either require manual fitting or manual spectral processing,
or a bit of both; see Appendix~I for a comprehensive list of \NMR\ softwares and their limitations.
The need for such manual interventions leads to a number of issues, including slower throughput, operator fatigue and associated operator errors,
the need for highly trained and dedicated experts, the requirement of two or more spectral assessments for quality assessment and control purposes, and inconsistent results between individuals, between labs or over different time periods~\cite{qnmr,between_person}. 

It would be better to have
a software system that can automatically perform both spectral processing and spectral deconvolution,
be 
able to analyze complex mixtures ($\sim$60 compounds)
quickly and accurately, and be able to produce reliable compound concentrations.  
Here we 
describe such a system, 
called 
\bayesil. 

Extensive testing,
on computer-generated and laboratory-generated chemical mixtures as well as real biological samples,
shows that \bayesil\  
consistently 
performs with $\sim 90\%$ accuracy for compound identification in mixtures with up to $60$ different metabolites.  It also determines metabolite concentrations with $\sim 10\%$ error. 
For computer-generated biofluid spectra, where the ground truth is known, \bayesil\ consistently outperforms highly trained experts in both identification and quantification.
\bayesil\ appears to be the first system that 
 supports 
fully automated and fully quantitative \NMR-based metabolomics. 
This paper describes this system, its underlying algorithms and
its performance across various tests.

\section{Metabolic profiling pipeline}\label{sec:method}
\bayesil\ performs 
fully automated spectral processing and spectral profiling for 1D \hone~\NMR\ spectra collected on either Agilent/Varian or Bruker instruments,
at several different frequencies.
In particular, it uses a variety of intelligent phasing and baseline correction methods to automatically process raw 1D \NMR\ spectra (\ie FIDs).  
It also uses approximate inference techniques to rapidly perform very accurate spectral deconvolution, yielding both compound identities and their concentrations. 
Here
we briefly describe \BAYESIL's spectral processing algorithms, the principles and rationale behind \BAYESIL's spectral deconvolution method and 
the construction of \BAYESIL's spectral library.

\comment{
\rg{Is "deconvolution" standard? Why not profilng?}\sr{David has used profiling to refer to  spectral (pre)processing + spectral deconvolution. Throughout the paper these terms are consistently used in this sense. }
}

\subsection{Spectral Processing in \small{BAYESIL}}\label{sec:preprocessing} 
 Successful \NMR\ spectral profiling depends critically on the quality and uniformity of the starting \NMR\ spectrum. 
Unfortunately, most spectral processing functions (\ie phasing, baseline correction, solvent filtering, chemical shift referencing) are left to the user. 
Given the complexity and 
large number of 
variables, values and filters that can be used, 
many view spectral processing more as an art, rather than a science.  
Different perspectives or different personal thresholds on what is a ``good looking'' \NMR\ spectrum can potentially lead to very different results
 regarding what compounds are identified or which compounds are accurately quantified in a biofluid spectrum. 
To address this issue, \bayesil\ 
itself performs all of the  spectral processing functions:
starting from the FID, it performs zero-filling, Fourier and Hilbert transformation, phasing, baseline correction, chemical shift referencing, reference deconvolution and smoothing.
Automating this process ensures reproducibility, consistency and uniformity of the input data prior to spectral deconvolution. Here we briefly sketch some of the more challenging steps in this process. 

\comment{\emph{Zero filling} involves extending the time-domain (FID) spectrum with zero values. 
This often produces higher resolution in the frequency domain. 
After zero filling and \emph{Fourier transformation},
all subsequent spectral processing is performed in frequency domain.} 

\emph{Phasing} involves  
maximizing the symmetry of the peaks by 
reducing zero-order and first-order phase mismatch.
Zero-order phase mismatch is a
sign of the difference between the reference phase and the receiver phase and 
is independent of frequency. 
The first-order phase mismatch can be a result of 
 the time-delay between excitation and detection, 
flip-angle variation and the filter that is used to reduce the noise outside of the spectral bandwidth~\cite{phasing_other}.
In addition to using well-known techniques, such as spectral norm minimization \cite{phasing_methods}, \bayesil\ uses the cross entropy optimization method~\cite{ceedthesis} to jointly maximize a direct measure of peak symmetry for isolated peaks across the spectrum. 

\emph{Baseline correction} involves removing distortions that may arise from hardware artifacts or highly concentrated components of 
the mixture (\eg solvent), while keeping the desirable signal intact. 
This process is often performed in two steps: 1) baseline-detection and 2) modelling.
\bayesil\ relies on iterative thresholding \cite{baseline_iterative} and estimating the
signal-to-noise ratio to detect the baseline points. It uses 
monotonic cubic Hermite interpolation \cite{pchip} and Whittaker smoothing technique for
baseline modelling \cite{whittaker}.

\bayesil\ also provides the options for \emph{smoothing and line-broadening} using Savitzky-Golay \cite{golay} and Gaussian filters.
However smoothing is mostly cosmetic and it is not essential for spectral deconvolution. 
In fact,
it may degrade the signal and occasionally remove the the low-amplitude and narrow peaks.
Similarly, we found 
the effect of \emph{reference deconvolution} -- which may be used to remove instrumental or experimentally induced distortions of the Lorentzian lineshape -- 
is also mostly cosmetic, and if the distortion around the reference peak has any source other than poor shimming, using reference deconvolution 
will have an adverse effect on the rest of the \NMR\ spectrum.

\subsection{Spectral Deconvolution}\label{sec:deconvolution}
An \NMR\ spectrum for a
 compound $\compound$
is a collection of one or more \emph{Lorentzian} peaks
formed into one or more clusters --
that is, each compound $\compound$ is a set of clusters $\{\cluster_k\}$,
where each cluster $\cluster_k$ is set of peaks,
and each peak is defined by a triple, 
$\pparams = (\pparam_1, \pparam_2, \pparam_3)$ 
corresponding to its height, center and width (at half height) respectively
-- where the height at $\y$ due to this peak is $\peak(\y; \pparams) = \frac{\pparam_1 \pparam_3}{\pparam_3 + 4 ( \pparam_2 - \y)^2}$.
Letting $\Y$ refer to the entire spectrum
(\eg from -1 to 13 \ppm\ when referenced against the \textsc{DSS} peak),
the height of the spectrum of a pure compound $\compound$
at each location $\y \in \Y$,  is 
\begin{equation} 
\label{eq:signature}
\predSpec{\compound,\conc_\compound,\shifts_\compound}{\y}
 \quad = 
\quad 
\conc_\compound
\sum_{\cluster \in \compound}\ \sum_{\pparams \in \cluster} 
  \peak(\y - \shift_\cluster; \pparams ) 
\end{equation}
where $\conc_\compound$ is the concentration of this compound
and $\shift_\compound = 
\{\shift_{\cluster}\, |\, \cluster \in \compound\}$ 
is the set of chemical shifts for the clusters 
associated with this compound.

An \NMR\ spectrum is essentially a linear combination of the peaks in its component compounds: 
that is, the height at each \PPM\ value $\y$ of a mixture spectrum is just the sum of the contributions of each compound.
This means, given the concentrations of the compounds 
$\concs = \{\conc_\compound \}$, and the chemical shifts
$\shifts = \bigcup_\compound \shifts_{\compound}$
of the clusters associated with these compounds, 
we can then ``draw'' an \NMR\ spectrum -- 
\ie the height at each \ppm\ $\y$, 
given this \concs\ and \shifts, is 
$\predSpec{\concs,\shifts}{\y}\ =\ 
\sum_{\compound} 
\predSpec{\compound,\conc_\compound,\shifts_\compound}{\y}$.

The spectral deconvolution challenge, in general, is the reverse process:
Given a set of compounds $\{\compound_1,\ldots,\compound_r\}$ with associated signatures
(\ie $\pparams$ values of their peaks, organized in clusters) and 
the observed spectrum $\spec(\cdot)$,
find the ``best'' combination of concentrations $\concs$ and shifts $\shifts$ to fit that spectrum.
To determine which values are best, for now, we consider a simple loss function that is the 
(square of the) difference
of the heights between the observed spectrum $\spec(\cdot)$ and the reconstructed spectrum
$\predSpec{\shifts, \concs}{\cdot}$\\[-0.2in]
\begin{align}\label{eq:loss}
\loss_{\Y}\left(\ \spec(\cdot),\ \predSpec{\concs,\shifts}{\cdot}\ \right) \; =\;
\int_{\y \in \Y} \big{(}\,\spec(\y)-\predSpec{\concs,\shifts}{\y}\,\big{)}^2 \, \mathrm{d}\y  
\end{align}
where the subscript $\Y$ indicates that this loss function applies to the entire spectrum; see Appendix~II for \bayesil's actual loss function.

Our task is to find the values of 
\begin{align} \label{eq:optimization}
[\concs^*,\ \shifts^*] \quad = \quad \arg_{\concs, \shifts}\min 
\quad \loss_{\Y}\left(\spec(\cdot),\, \predSpec{\concs,\shifts}{\cdot}\right)
\end{align}
that minimize this loss function.
\refFigure{fig:fit} shows part of a spectrum over a complex mixture, and \bayesil's
solution obtained by minimizing the loss function.

This corresponds to search over a huge space -- all possible shifts 
for each of the clusters, and all possible concentrations over the compounds.
The key innovation of \bayesil\ is how it minimizes this highly non-linear loss function, efficiently. 
In particular, \bayesil\ ``factors'' this large task into a set of inter-related smaller tasks.
Two characteristics of the \NMR\ spectra make this factorization possible:
1) each shift is over only a small range
(typically a window of $\pm 0.025$ \PPM); 
and
2) as the height in the spectrum due to a Lorentzian peak diminishes quadratically from its center,
each peak and therefore each cluster can only ``influence'' a small interval;
see Appendix~III.

Now consider a function that maps each point in the spectrum to the set of clusters that might affect the height at this location.
\comment{
That is, given the set of compounds $\compounds(B)$ that might appear in a particular 
biofluid B (\eg the 48 that can appear in CSF),
we can identify each point $\y \in \Y$ with the small set of clusters that might influence it: $\clusters_{B}(\y) \subset \compounds(B)$.
}
That is, given the set of compounds $\{\compounds\}$ that might appear in a particular 
biofluid (\eg the 48 that can appear in CSF),
we can identify each \PPM\ location $\y \in \Y$ with the small set of clusters that might influence it, $\clusters(\y)$.

We can then partition the spectrum into disjoint contiguous regions,
$\{ \Y_I \}$,
where every \PPM\ location 
in each $\Y_I$ involves exactly the same subset of clusters
-- \ie for any pair of points $\y_1, \y_2 \in \Y_I$, we know that $\clusters(\y_1) = \clusters(\y_2)$.
For example, our library for CSF includes 48 compounds, with a total of 180 clusters and  946 peaks. 
A typical CSF spectrum is partitioned into  $\sim$350 regions. 
Each of these regions will involve between 1 and $\sim$25 clusters, 
and span between 0.0001 and 1.5~\PPM. 
Moreover, each cluster will appear in between 1 and $\sim$70 different regions.

\begin{figure}
\includegraphics[width=1\linewidth]{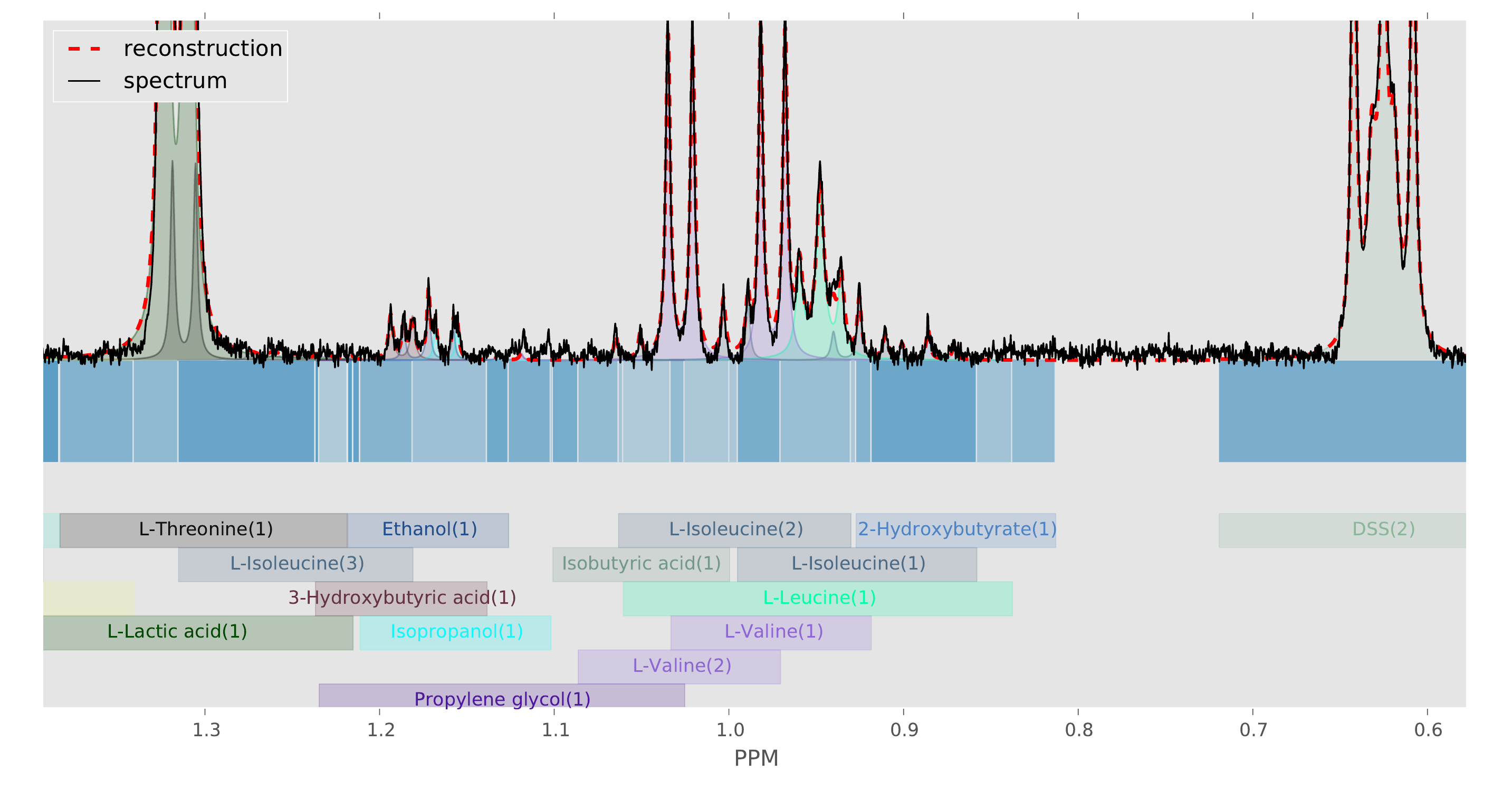}
\caption{Partitioning of $\Y$ into continuous blocks $\Y_{\I} \subset \Y$ for a part of a human serum spectrum.
Here each block is shown with a different shade of blue, below the horizontal axis.
The domain of influence of each cluster is also indicated with coloured blocks.} 
\label{fig:blocks}
\end{figure}

\refFigure{fig:blocks} 
shows the division of a part of human serum \NMR\ spectrum into
regions $\Y_I$; blocks in different shades of blue.
For example the region $\Y_{[.8563, 9289]}$ from 0.8563 to 0.9289 \PPM\ might include significant contributions from the first cluster of 2-Hydroxybutyrate, the first cluster of L-Isoleucine 
and/or the first cluster of L-Leucine.
The region immediately to its left (from 0.9289 to 0.9370 \ppm)
includes these and also a cluster of L-Valine,
and the one to the right (from 0.8563 to 0.87526 \ppm) 
does not include L-Isoleucine.


As the loss function $\loss(\cdot,\cdot)$ is additive over the domain $\Y$, 
we can rewrite the optimization of \refEq{eq:optimization} 
as the sum of the losses for each of the regions $\Y_I$:
\begin{align}\label{eq:decompose}
[\concs^*,\shifts^*] \; = \; \arg_{\concs,\shifts}
\min \sum_{I} \loss_{\Y_\I}\big{(}\,\spec(\cdot), \predSpec{\concs_I, \shifts_I}{\cdot}\,\big{)}
\end{align}
\\[-1ex]
Now recall that each region $\Y_I$ involves relatively few compounds and clusters.
This suggests a preliminary step of simply ``solving'' each region, by itself: \ie find the best centers for the clusters in that region $\shifts_I$, and the best concentrations for the associated compounds $\concs_I$, which collectively minimize the loss over the \PPM-interval $\Y_I$.
This simple approach is fast, as it involves relatively few variables
and 
a limited range of \PPM-values.
Unfortunately, this does not produce the overall correct answer
-- that is, each region has an opinion about the concentration and shift values of its cluster, 
and when two (or more) regions each involve the same variable, they must both agree on its value. 



\begin{figure}
\includegraphics[width=1\linewidth]{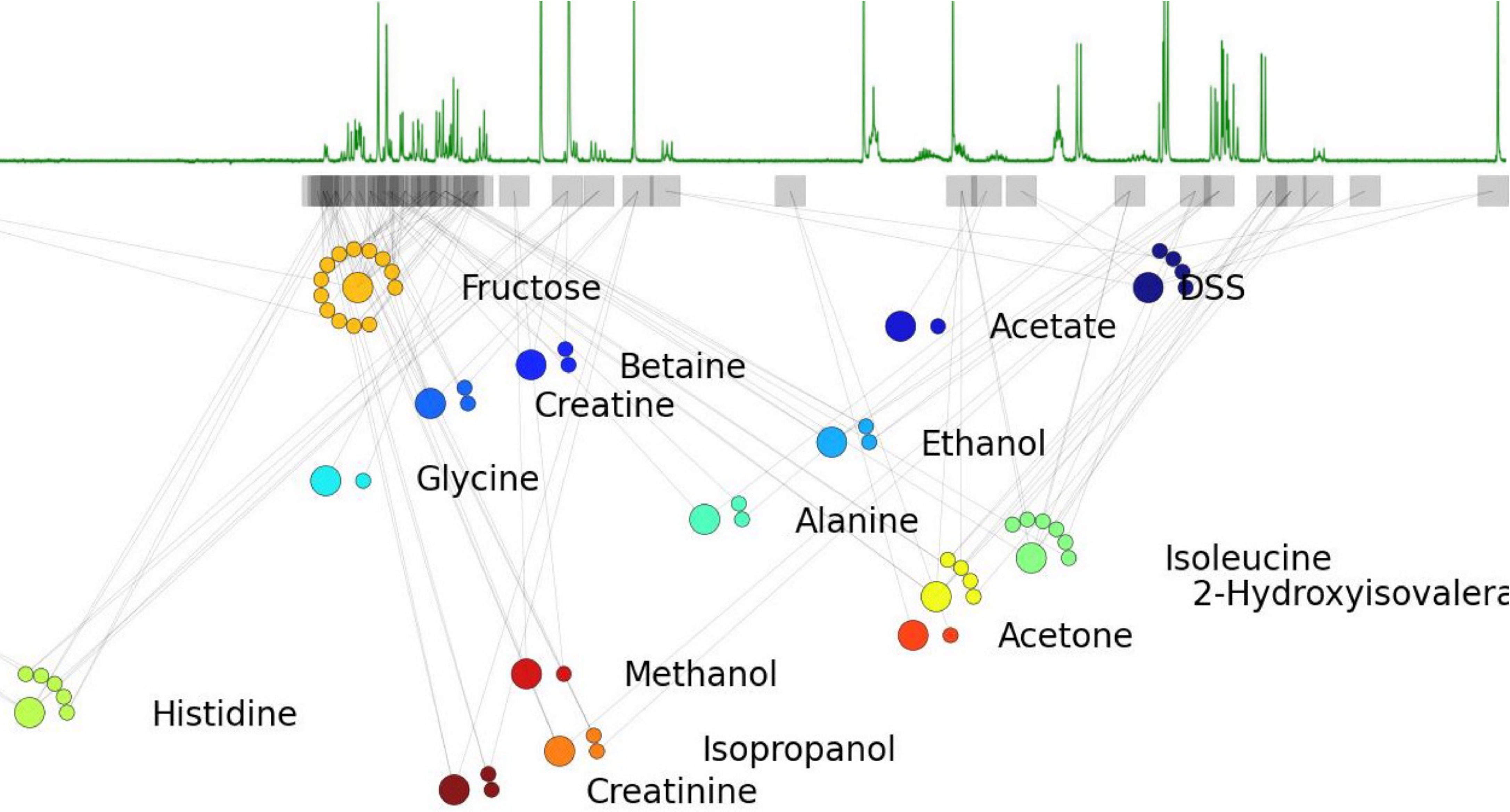}
\caption{Factor-graph for a library of 15 compounds 
immediately below 
an associated \NMR\ spectrum.
Each factor is represented by a square and each variable using a circle. Concentration (larger circles) and shift variables (smaller circles, beside the associated concentration)
corresponding to each compound appear together. 
The position of each factor $\ff_{I}$ position in the plot corresponds
to the center of the corresponding block $\Y_I$.}
\label{fig:pgm}
\end{figure}

To address this problem, we take a probabilistic approach, viewing the task of minimizing the loss function \refEq{eq:loss}
as finding the ``Maximum a Posteriori'' (MAP) assignment
-- \ie the assignment to all of the 
 cluster-shift and compound-concentration $[\shifts, \concs]$
variables that makes the observed data as likely as possible.
Here, the Boltzmann formula gives the probabilistic interpretation of the loss 
(\aka\ the energy) 
\begin{align}\label{eq:boltzmann}
\pp(\,\concs,\shifts\,|\, \spec(\cdot)\,) = \frac{1}{Z} \exp\left\{-\frac{1}{T}\loss_{\Y}\big{(}\,\spec(\cdot),\, \predSpec{\concs, \shifts}{\cdot}\,\big{)}\right\} 
\end{align}
where $Z$ is the normalization constant and $T$ is known as the ``temperature''  parameter,
explained in Appendix~IV. 
Using the decomposition of loss over regions (\refEq{eq:decompose}) we can write this distribution
in factored form
 \begin{align}  
\pp(\,\concs,\shifts\,|\, \spec(\cdot)\,) &= \frac{1}{Z} \prod_I \ff_I(\concs_I,\shifts_I) \nonumber\\
\ff_I(\concs_I,\shifts_I) &= \expp{-\frac{1}{T}\loss_{\Y_I}\big{(}\,\spec(\cdot),\, \recspec(.; \concs_I, \shifts_I)\,\big{)}}
\label{eq:factored}
\end{align}
\noindent 
This decomposition of the distribution $\pp(\,\concs,\shifts\,|\, \spec(\cdot)\,)$ 
can be represented using a probabilistic graphical model, 
known as a factor graph~\cite{pgm}, 
which 
is a graph with 
two types of nodes: 
1) factors (corresponding to regions or $\ff_I$), and 2) variables (here, concentrations and chemical shifts). 
Each factor has arcs that point only to its associated variables.
\refFigure{fig:pgm} shows a portion of this factor-graph for a simple defined mixture of 15 compounds.

\begin{figure}
 \includegraphics[width=1\linewidth]{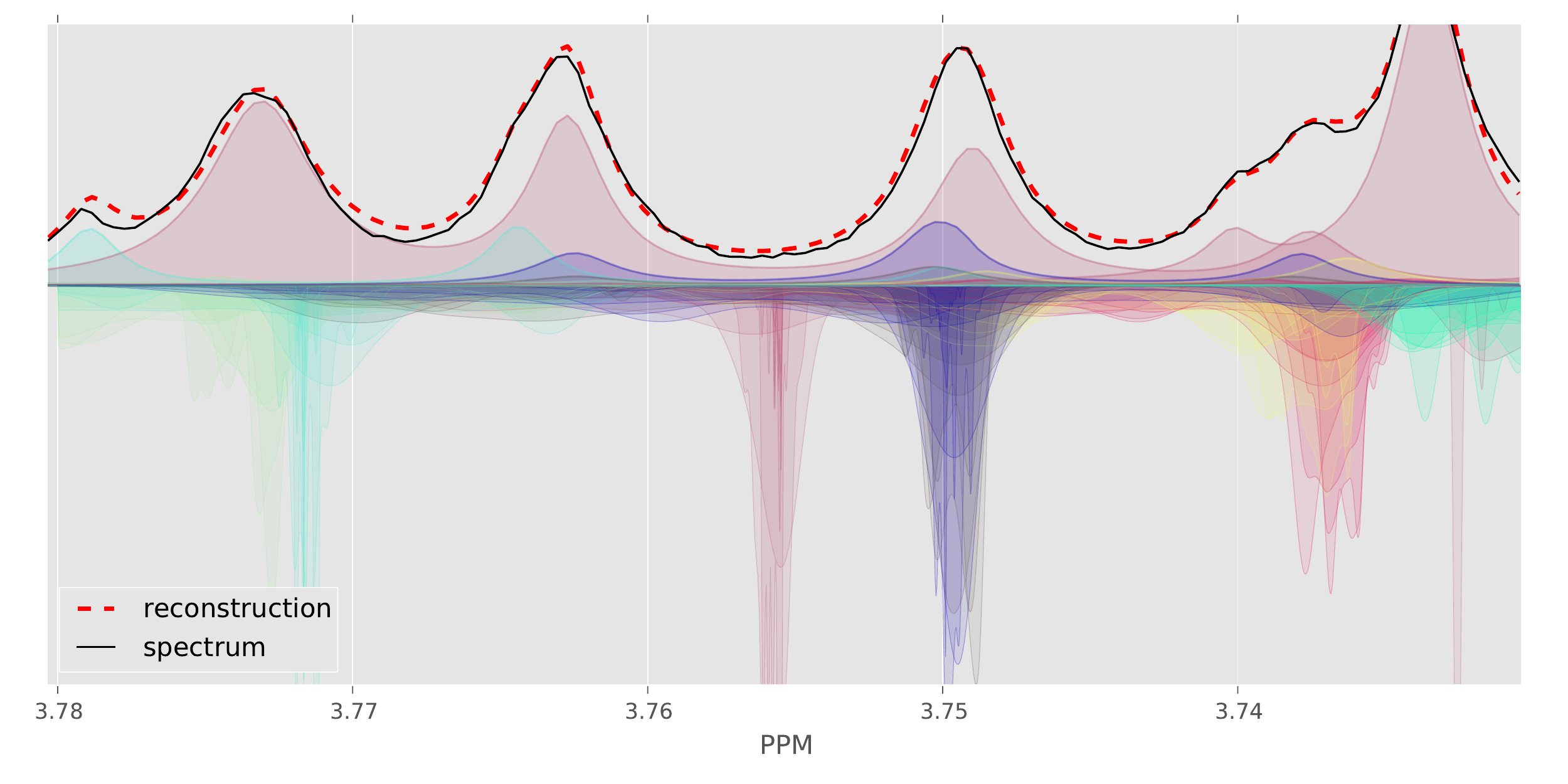}
\caption{A small region of human serum spectrum.
The plots above horizontal axis show the original spectrum (solid black), individual clusters as well as overall fit (dashed red).
The curves below horizontal axis show the \bayesil's distribution over chemical shift variables  for each cluster ($\cluster$), 
over $6$ iterations of spectral deconvolution. The distributions become more peaked towards the correct center in each iteration. 
Distributions below the horizon have the color of their associated cluster.
}
\label{fig:kde}
\end{figure}
By formulating the spectral deconvolution problem as MAP inference in a factor-graph,
we have a variety of inference techniques at our disposal~\cite{pgm}. 
\bayesil\ uses 
a non-parametric sequential Monte Carlo method~\cite{sequentialmontecarlo} tailored to  
this inference problem 
(see Appendix~IV for details).
Using a Gaussian distribution around a set of particles (\aka~kernel density estimation) \bayesil\ represents a distribution over each concentration 
$\conc_\compound$ and shift variable $\shift_\cluster$.
By reducing the temperature parameter, these distributions are gradually narrowed in each iteration until convergence, at which point the mode of the distributions represent an approximate MAP assignment.
\refFigure{fig:kde} shows the evolution of distributions over the chemical shift variables over $6$ iterations of spectral deconvolution.

\comment{
The concentrations that we obtain at this stage (\ie $\concs$) are relative.
\bayesil\ uses a reference compound (\eg 4,4-dimethyl-4-silapentane-1-sulfonic acid, \aka DSS) with known concentration, to obtain the absolute quantities.
}

\subsection{\small{BAYESIL}'s Spectral Library}\label{sec:library}
Key to the success of any spectral deconvolution algorithm is the quality and size of its spectral library. 
We therefore collected 1D \hone~\NMR\ reference spectra for each of the 
compounds in \bayesil's spectral library using pure compounds obtained from the 
Human Metabolome Library~\cite{hmdb},
using a standard protocol (see Materials and Methods).
\comment{
After spectral collection, the spectra were checked for quality and then analyzed using a locally developed spectral analysis tool 
to convert the spectra into a series of XML files,
}%
The spectral library 
contains relevant information about each compound ($\compound$) including individual peak clusters ($\cluster$) and peak amplitude positions and widths (\ie $\pparams$ in \refEq{eq:signature}), 
as well as allowable chemical shift window $\underline{\shift}_\cluster \leq \shift_\cluster \leq \overline{\shift}_\cluster$ for each cluster $\cluster$.

To analyze each biofluid, \bayesil\ uses
a specific spectral sub-library -- here, one for serum and another one for CSF.
The serum library consists of 50 \NMR-detectable compounds from the human serum metabolome~\cite{metabolome_serum} 
while the CSF library consists of the 48 \NMR-detectable compounds from the human CSF metabolome~\cite{metabolome_csf}; see Appendix V.
The use of biofluid-specific or organism-specific spectral libraries significantly improves the performance of the spectral fitting process
as it reduces the number of possible explanations for each peak.

\section{Assessment}\label{sec:assessment}
\BAYESIL\ was assessed using 3 different types of spectral data sets over two different types of biofluids:\\
\textbf{(a) Computer generated mixtures derived from its spectral library.}
We generated 5 random serum 
and 5 random CSF
spectra by sampling from the distribution of the measured concentration ranges of various compounds,
 and the probability of observing them in the mixture
from~\cite{metabolome_serum,metabolome_csf}. 
The chemical shifts were also randomly sampled according to the  
chemical shift ranges from the corresponding spectral libraries.
To be more realistic, we also added a small amount of noise to the spectrum -- independently to the height at each position $\y \in \Y$.
These correspond to ``perfect'' spectra,
and are intended to assess the robustness and performance limits of \BAYESIL\ under optimal conditions.\\
\textbf{(b) Defined mixtures prepared in the laboratory.}
We created 15 defined mixtures (5 defined mixture of serum, 5 defined mixture of CSF, 5 random mixture of compounds in both serum and CSF,  involving $>60$ compounds), 
using carefully measured pure compounds and freshly prepared solutions.
These provide real spectral data that probably include common spectral and solution artifacts (baseline and phasing issues, minor spontaneous reaction products, contaminants, matrix or pH effects). 
This set was used to assess \BAYESIL's performance under well-controlled conditions where the composition and of the mixtures was almost perfectly known.\\
\textbf{(c) Biological serum and CSF samples.}
We took human CSF and serum samples from previously studied samples that had been analyzed and quantified by \NMR\ experts
-- here, 
50 human serum and 5 human CSF samples.
The set of compound mixtures was used to assess \BAYESIL's performance under realistic conditions with common spectral 
and solution artifacts.  
Although  human CSF contains a smaller number of 
\NMR-detectable compounds than human serum,  it is more difficult to profile due to the lower concentration of metabolites.
While both the biological samples and defined mixtures were thoroughly analyzed, their exact compound concentrations cannot be perfectly known.\\
Overall, we believe these 3 test sets provide a robust assessment of \BAYESIL's performance 
(as well as its 
limitations) under a wide range of conditions.

\comment{ merged in above:
The computer-generated set {\textbf (a)} corresponds to ``perfect'' spectra
and was intended to assess the robustness and performance limits of \BAYESIL\ under optimal and perfectly controlled conditions. 
The defined mixture set {\textbf(b)} corresponds to real spectral data that would likely include spectral and solution artifacts (baseline and phasing issues, minor spontaneous reaction products, contaminants, matrix or pH effects). 
This second set was intended to assess \BAYESIL's performance under well-controlled conditions where the composition and of the mixtures was almost perfectly known.  
The third set of compound mixtures correspond to real biofluid data that would also be expected to contain spectral and solution artifacts.  
Although the human CSF contains smaller number of \NMR-detectable compounds than human serum, 
due to lower concentration of metabolites, it is more difficult to profile.
While both the biological samples and defined mixtures were thoroughly analyzed, their exact compound concentrations cannot be perfectly known.  
Overall, these 3 test sets were selected to provide a robust assessment of \BAYESIL's performance (as wells as its limits and limitations) under a range of conditions.
}

\begin{table*}
  \caption{Identification and quantification accuracy of \bayesil\ and human expert on various data-sets. 
}\label{table:accuracy}
\centering
\begin{tikzpicture}
\node (table) [inner sep=0pt] {

    \begin{tabu}{c  l |[2pt] r  c  l |[2pt] r  c  l|[2pt] c }
      \multicolumn{2}{c|[2pt]}{}&\multicolumn{3}{ c |[2pt]}{serum}& \multicolumn{3}{ c|[2pt] }{CSF}&\multicolumn{1}{ c }{complex}\\
      \multicolumn{2}{c|[2pt]}{}&
      biological&
      def. mix.&
      comp. gen.&
      biological&
      def. mix.&
      comp. gen.&
      def. mix. 
      \\\tabucline[1pt]{3-9}
      \multirow{2}{*}{\bayesil }&
id. accuracy & $.93\pm.04$ & $.94\pm.02$ & $.98\pm.01$ & $.90\pm.04$ & $.89\pm.03$ & $.95\pm.03$ & $.90\pm.02$\\
&quant. accuracy & $.89\pm.02$ & $.90\pm.02$ & $.98\pm.01$ & $.91\pm.01$ & $.90\pm.02$ & $.94\pm.02$ & $.88\pm.02$ \\\tabucline[1pt]{1-9}
      \multirow{2}{*}{ expert  }&
      id. accuracy & - & - & $.91\pm.02$ & - &- & $.87\pm.05$ & -\\
      &quant. accuracy & - & - & $.95\pm.01$ & - & - & $.91\pm.04$ & -\\
\end{tabu}
};
\draw [rounded corners=.5em] (table.north west) rectangle (table.south east);
\end{tikzpicture}
\end{table*}
\comment{
              dataset id.mean quant.mean id.sd quant.sd
1    biological_serum   0.929      0.891 0.037    0.016
2       def_mix_serum   0.943      0.898 0.021    0.020
3      biological_csf   0.899      0.914 0.036    0.014
4         def_mix_csf   0.893      0.899 0.027    0.016
5 def_mix_serum_n_csf   0.952      0.883 0.033    0.022
}

\bayesil\ estimates the ``detection threshold'' based on the signal to noise ratio (\textsc{snr})
in each spectrum -- \ie when the signal is noisy this threshold is increase to provide a more
confident identification and quantification. 
The \textsc{snr} and therefore the detection threshold is directly
related to the number of scans during spectral acquisition. For example our biological serum
samples are produced using $128$ scans and therefore most detection thresholds are $\sim 10 \mu$M. Since compound concentrations in our CSF samples are considerably lower than serum, their analysis requires higher quality spectra (see Materials and Methods). Our CSF samples use $1024$ scans resulting in detection thresholds that are often  less than $2 \mu$M.
However this threshold is not uniform across metabolites.
\bayesil\ also uses a relative factor in compound detectability; as some compound such as Choline are
easy to identify and quantify at low concentrations while for some other compounds such as L-Asparagine, experts use a
 higher detection threshold.

Given a spectrum of a mixture of compounds (with ``true'' concentrations $\{{\conc}_{\compound}\}$), 
\bayesil\ returns its {\em estimates}\ of these concentrations $\{\widehat{\conc}_{\compound}\}$, 
which might be 0 if that compound is absent.
We say a compound is a true positive 
if both $\widehat{\conc}_{\compound}$ and $\conc_{\compound}$ are 
positive -- that is, greater than the detection threshold,
and
a true negative if both $\widehat{\conc}_{\compound}$ and $\conc_{\compound}$ are
less than the threshold;
in either case,
\bayesil's prediction is considered correct.
\bayesil's identification accuracy for a given spectrum is the ratio of correct labels 
(true positives plus true negatives) to the library size.
\bayesil's ``quantitative accuracy'' describes how often its estimates  
$\widehat{\conc}_{\compound}$ were 
``close enough'' to the true values $\conc_{\compound}$;
note that simply computing $|\widehat{\conc}_{\compound} - {\conc}_{\compound}|$
is not enough as this measure would basically
only consider the compounds with high concentrations.
We instead use the 
$\mathrm{median}_{\compound} \left(
  \frac{|\conc_\compound - \widehat{\conc}_\compound|}{\max(\widehat{\conc}_\compound, \conc_\compound)} 
   \right)$
as a measure of the percentage error in concentrations.
\refTable{table:accuracy} reports \bayesil's identification and quantification accuracies,
for each of the tasks listed above.
For the biological and synthetic samples, we assume the human expert's assessment is correct.
\comment{
  For the computer generated spectra {\bf (a)}, the exact ground truth $\{ \conc_\compound\}$
is known. 
For  synthetic {\bf (b)} and biological {\bf (c)} mixtures,
we used the expert's estimate of the concentrations as ground truth.
}%


\begin{figure} 
\centering
  \includegraphics[width=\linewidth]{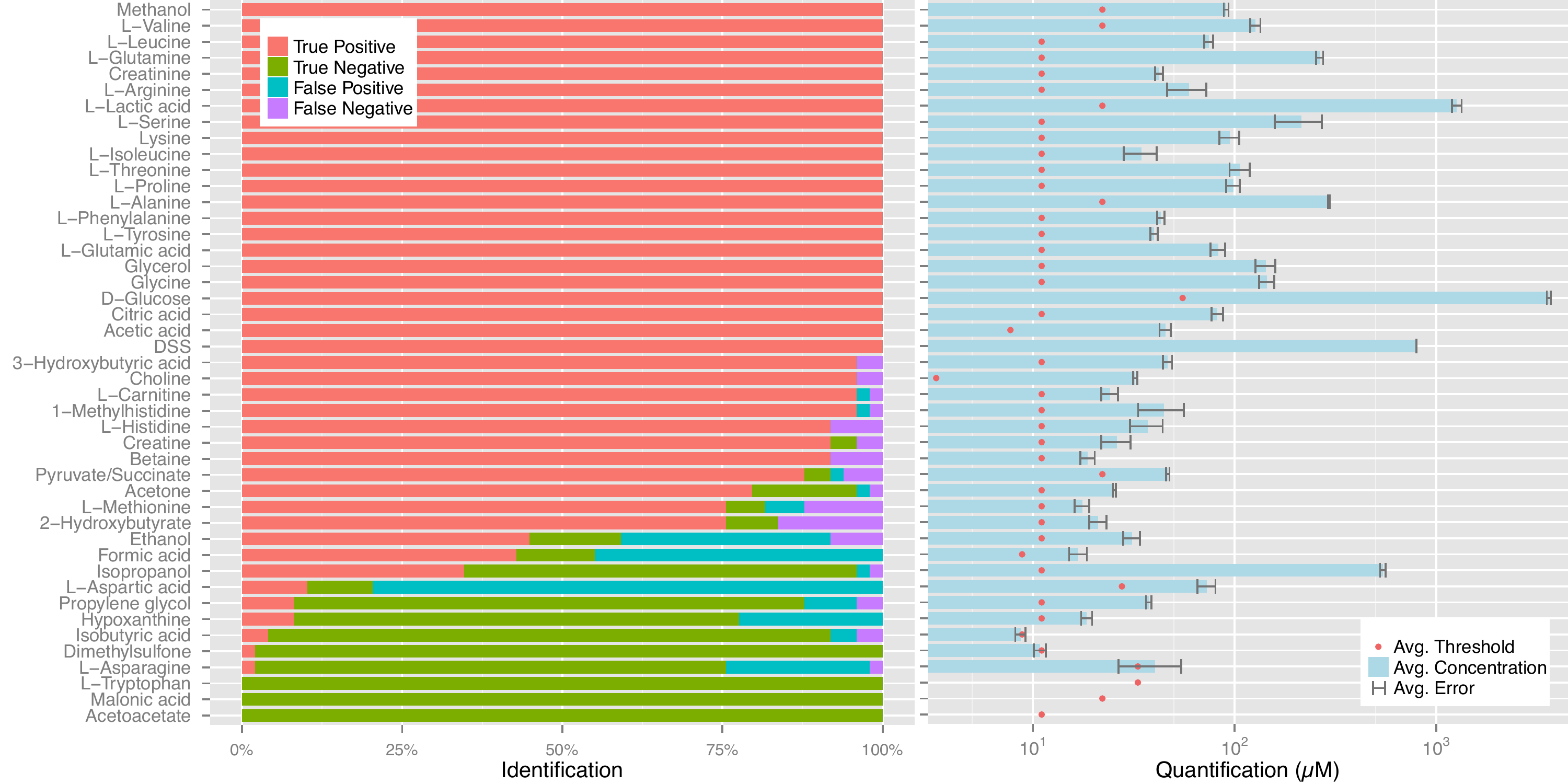}
\caption{
(left) \bayesil's true/false positive/negative rate in identification of individual compounds in 50 biological serum samples.
(right) The average concentration for correctly identified compounds in the same samples.
The error bars show the average difference between \bayesil\ and expert values for each compound and the red dots show the detection threshold for each compound.}
\label{fig:compound_error}
\end{figure}

\refFigure{fig:compound_error}(left) reports the frequency of false/true positives/negatives 
for individual compounds in 50 serum samples.
\refFigure{fig:compound_error}(right)
shows the average of $\conc_\compound$ for correctly identified compounds in 50 serum samples, as reported by \NMR\ experts, the average detection threshold for diffent compounds 
as well as the average difference $\widehat{\conc}_{\compound} - \conc_{\compound}$,
between \bayesil\ and expert's estimate for each compound.

These results on a diverse set of test data suggest that \bayesil\ is often within $10\%$ of the expert's estimate, 
and where the ground truth is known,
\bayesil's metabolic profile is often more accurate than the expert's.
\bayesil's web-page \textit{http://www.bayesil.ca} provides a complete description of all of the studies reported above,
showing the fits and the metabolic profiles obtained. 


\subsection{Efficiency}
On a single 2.8 GHz CPU processor, \bayesil\ typically takes less than 5 minutes to  profile a serum or CSF spectrum. 
Over a sustained 24 hour period, \BAYESIL\ should be able to process more than 200 spectra
(vs.~$\sim$20 spectra/day for a human expert) 
and accurately identify-\&-quantify approximately 50 compounds per spectrum.  
This corresponds to an output of more than 5000 metabolite measurements a day for a single CPU.  
It is possible to produce profiles yet faster by reducing the number of particles used in \BAYESIL's deconvolution process:
reducing this by a factor of 10, reduces the run-time to less than one minute.
This speed-up often reduces the identification and quantification accuracy by only 5-10\%.
\subsection{Limitations}
A key disadvantage of \NMR\ analysis (versus mass spec), 
in general, 
is that \NMR\ is relatively insensitive, 
with a lower limit of detection of $\sim 1 -5$ \uM\ and a requirement of relatively large sample sizes ($\sim 500 \mu L$).
For high-quality spectra with high \textsc{snr}, such as CSF samples in our study, \bayesil\ 
is often able to achieve its high accuracy within these detection limits.

Key to the high level of performance of \BAYESIL\ is the use of biofluid-specific spectral libraries in its spectral fitting routines (\aka targeted profiling). 
Without these, \BAYESIL\ would be 
substantially slower and less accurate in both identification and quantification.
Therefore users must provide \BAYESIL\ with information about the biofluid being analyzed.
Obviously, the mis-identification or mislabelling of a biofluid sample could lead to somewhat poorer 
results. 

\comment{As \bayesil\ is doing ``targeted profiling'', 
its spectral fitting routine requires a spectral library.
In fact, we use libraries that are specific to the biofluid being analyzed;
this allows \bayesil\ to be substantially faster and more accurate in both identification and quantification.}


This need for prior knowledge about the typical composition of biofluid mixtures
has motivated us, and others, to spend considerable efforts 
to determine the \NMR-detectable metabolomes for 
many biofluids, including human plasma/serum, 
cerebrospinal fluid, 
saliva and urine (\eg \cite{metabolome_serum,metabolome_csf,metabolome_urine}), 
milk and rumen (\eg \cite{metabolome_milk}), 
cell extracts (\eg \cite{metabolome_mammalian}), 
cancer cells \cite{metabolome_cancer,metabolome_cancer2},
and many other fluids or extracts.



\subsection{Other systems}
There have been a number of software packages 
recently
developed to facilitate spectral profiling and compound identification/quantification by \NMR; 
see Appendix~I for details.
However, they do not seem to be particularly accurate on real biofluids. 
Furthermore, many of the packages are not currently publicly available.
To date, the largest number of metabolites that has been automatically identified and quantified using publicly available software is 26 compounds, by \BATMAN~\cite{batman}. 
However, an analysis of this magnitude required several hours of CPU time to process a single spectrum.  
Furthermore, \BATMAN\ also requires a human expert to perform many of the preliminary steps.
We compared \BAYESIL\ to \BATMAN\ on simple computer-generated mixtures, involving 5, 10 and 20 compounds selected from \BATMAN's library, as well as a preprocessed human serum sample. For the computer generated spectra, both \batman\ and \bayesil\ used identical libraries containing only the relevant compounds. \batman\ achieved $85-87\%$ quantification accuracy for the  computer-generated mixtures but took 2-9 hours to run, while in all cases \bayesil\ achieved $>98\%$ accuracy in less than 3 minutes. For the serum spectra, \bayesil\ used a library of 50 compounds, while \batman\ used a subset of 40 compounds that its library has in common with the serum metabolome. \batman\ took 19 hours to analyze the serum spectrum and identified all the compounds in its library (resulting in $85\%$ identification accuracy) and only $8\%$ quantification accuracy compared to \bayesil\, which took 5 minutes to achieve $98\%$ identification and $90\%$ quantification accuracy. 


\section*{Conclusion}
\NMR\ is a  particularly appealing platform for conducting metabolomic studies on biofluids 
as it is a rapid, robust, highly reproducible, non-destructive, and fully quantitative technique that 
requires no prior compound separation or derivatization. 
The main barrier 
preventing widespread adoption of NMR-based metabolomics in industrial or clinical applications
is the requirement for manual spectral profiling.
\bayesil\ addresses this critical problem by providing fully automated spectral processing and deconvolution.
Furthermore it is able to perform this task on mixtures containing $>50$ compounds, 
with $90\%$ accuracy.
We believe that removing the automation barrier will 
have a significant, positive impact on NMR spectroscopy and \NMR-based metabolomics.
\BAYESIL\ is freely available for users to perform metabolic profiling of 1D \hone~\NMR\ spectra of serum, CSF and other biofluid mxitures (excluding urine) collected at 500 and 600 MHz.
\section{Materials and Methods}
{\small
To produce each of the reference spectra for \bayesil's library,
we first prepared stock solutions (1 mM to 100 mM) for each compound in 1 L in volumetric flasks.  
The metabolites were dissolved in 20 mM NaHPO$_4$ (pH 7.0). 
These stock solutions were further diluted if necessary to obtain a final stock solution concentration of 1 mM. The final sample for \NMR\ was prepared by transferring 1140 \uL\ to a 1.5 mL Eppendorf tube followed by the addition of 140 \uL\ D$_2$O and 120 \uL\ of the reference standard solution (11.67 mM DSS (disodium-2,2-dimethyl-2-silapentane-5-sulphonate), 20 mM NaHPO$_4$, pH 7.0).  
After confirming that 
the pH of the sample was between 6.8 and 7.2
(adjusting the buffer if necessary),
we transferred 700 \uL\ to a standard \NMR\ tube for spectral acquisition. 
All library \hone~\NMR\ spectra were collected on both 500 MHz and 600 MHz Inova spectrometers equipped with 5 mm Z-gradient PFG probes. A standard presaturation \hone-NOESY experiment(tnnoesy.c) was acquired at 25$^o$C 
using the first increment of the presaturation pulse sequence.  A 4 s acquisition time, a 100 ms mixing time, a 10 ms recyle delay and a 990 ms saturation delay were chosen.  Thirty-two transients were acquired for samples collected at 600 MHz while 128 transients were acquired for all samples collected at 500 MHz. Eight steady state scans were employed and the presaturation pulse power was calibrated to provide a field width no greater than 80 Hz.  Both the transmitter offset and the saturation pulse were centered on the water resonance and no suppression gradients were used.  After spectral collection, the spectra were checked for quality and then analyzed using a locally developed spectral analysis tool 
to convert the spectra into a series of XML files.
In producing the XML library, most peak clusters were given a default shift-window of 0.025 ppm, with the exception of few compounds such as histidine or citrate that are known to be highly pH-sensitive. 
Both the synthetic and real biological spectral data were collected in the manner described above except for biological CSF in which 1024 scans were collected to compensate for dilution.  For sample preparation, CSF was used as is, while serum was obtained after the blood had clotted for 30 min at 25$^o$C and then passed through pre-rinsed 3000 MWCO Amicon Ultra-0.5 filters to remove remaining proteins.  In each case 285 \uL\ of filtrate was obtained and 35 \uL\ of D$_2$O and 30 \uL\ of buffer was added. A total of 350 \uL\ was then transferred to a suitable Sigma tube for NMR data acquisition.  In the case of biological CSF, where less than 285 \uL\ was obtainable, the samples were diluted with sufficient H$_2$O.
}

\section*{Acknowledgments}
We gratefully acknowledge support from 
the Alberta Innovates -- Health Solutions,
the Alberta/Pfizer Translational Research Fund
and the Metabolomics Innovation Centre 
(funded by Genome Canada and Genome Alberta) 
for the project as a whole,
and to NSERC, CIHR, 
the Alberta Innovates Centre for Machine Learning 
(funded by Alberta Innovates -- Technology Futures), QEII and AIFT
graduate scholarships 
for individual funding.






\clearpage

\appendix[I. Other NMR-analysis software tools]
Several software packages have been developed to facilitate \NMR\ spectral processing, compound identification and quantification.  While some facilitate two dimensional \NMR\ spectral processing and compound identification (\eg \cite{colmar, metabominer}, dataChord (One Moon Scientific)),
none are completely automated
and provide compound quantification.
Furthermore, because modern metabolomic \NMR\ studies require the analysis of a tremendous amount of spectral data, the time required to collect 2D \NMR\ spectra makes their implementation prohibitive not only due to their inherent inability to facilitate high-throughput studies but also due to complicating factors such as sample degradation within the spectral acquisition time frame.  Thus, more effort has been directed at developing software tools 
for 1D \NMR.

Some tools provide basic processing functionalities~\cite{nmrglue,nmrpipe}.
(In fact, \bayesil\ uses \textsc{nmr-glue}~\cite{nmrglue} for handling different input/output data formats.)

We focus on software package capable of handling high-throughput 1D metabolomic \NMR\ data.
An ideal tool here should have the following features: 
(1)~{\em fully automated} -- in both spectral processing and compound identification and quantification; 
(2)~{\em flexible and customizable} -- capable of analyzing a wide (and extendable) range of different 
biological fluids;  
(3)~{\em ubiquitous} -- can accommodate input from different NMR vendors, multiple spectrometer frequencies; 
and of course
(4)~{\em accurate}.

Various software packages have made incremental steps toward achieving these goals.  
Of the 19 that we could identify,
only a handful provide some degree of automated identification and/or quantification (1):
\cite{colmar,bquant,hires,autofit,ceed2,batman2,newmethod}, Juice Screener, Wine Screener and Metabolic Profiler (Bruker Corporation) and Chenomx NMR Suite (Chenomx Inc.).
This task (of spectral deconvolution and metabolite profiling) has been tackled with a variety of algorithmic approaches
-- including
simple text file matching, binning~\cite{bquant},
principal component analysis and non-negative matrix factorization~\cite{colmar,hires}, combinations of simulated annealing and gradient descend (\cite{autofit2}, Chenomx NMR Suite (Chenomx Inc.)), cross entropy method~\cite{ceed2} and Monte Carlo techniques~\cite{batman2}. 
However only a few of these softwares provide automated spectral processing (\cite{lcmodel},Juice Screener and Wine Screener).

In terms of flexibility and customizability (2),
some software packages do utilize 
large data-sets (HMDB~\cite{hmdb2}, BMRB~\cite{bmrb} or MMCD~\cite{mmcd}) 
but they still require
the user to select a subset of the compounds,
and/or do not provide quantification~\cite{colmar,metabohunter,metabominer}.
Others are specialized to particular mixtures (\cite{lcmodel}, Wine Screener, Juice Screener
and Vantera (LipoScience Inc.)) and none can accurately quantify complex mixtures (with $> 50$ compounds).
Moreover, many of these software packages are specific to a particular
instrument  (\eg \cite{batman2,ceed,newmethod,crockford}, Wine Screener, Juice Screener, dataChord and Vantera).

It is often difficult to access accuracy (4), 
as the descriptions of many software tools do not provide any assessment
(\eg Juice Screener, Wine Screener and Metabolic Profiler 
(Bruker Corporation), \cite{colmar,lcmodel})
and many systems have been assessed merely on very simple mixtures (\eg~\cite{metabominer,metabohunter,batman2},lipoprofiler/Vantera (LipoScience Inc.)) or simple spike-in experiments~\cite{bquant,newmethod,crockford}.

To summarize, \bayesil\ is the only 1D 
\hone~\NMR\ interpretation system that is
completely automated (both preprocessing and deconvolution) for a wide range of complex mixtures (\ie all mammalian biofluids covered by its current library; see \textit{www.bayesil.ca}), involving $>60$ compounds. Moreover, it is efficient, general, accurate and publicly available.

\comment{
\rg{I do not understand the structure of this paragraph.
Can we systematically say:  Who did not do 1. Then who did not achieve 2. Then ... ??
Or maybe it is doing that, but I am not seeing it?}
These software packages also differ in their speed, their database size, 
the maximum number of analyzable metabolites or signals, 
whether they can quantify the metabolites and finally their accuracy with respect to both identification and quantification. 
\rg{Is next sentence just wrt accuracy?}
This is often different to evaluate as 
many software packages have not provided such assessments 
(\eg Juice Screener, Wine Screener and Metabolic Profiler 
(Bruker Corporation), \cite{colmar,lcmodel}) 
or have been assessed merely on very simple mixtures (\eg~\cite{metabominer,metabohunter,batman2},lipoprofiler/Vantera (LipoScience Inc.)) or simple spike-in experiments~\cite{bquant,newmethod,crockford}.
Some approaches utilize large data-sets (HMDB~\cite{hmdb2}, BMRB~\cite{bmrb} or MMCD~\cite{mmcd}) 
but may require 
\rg{is this correct:}
the user to select a subset of the compounds
to improve speed and accuracy, 
\rg{isn't that what Bayesil does??}
or will simply identify compounds in simple mixtures but not provide quantification~\cite{colmar,metabominer,metabohunter}. 
Others offer improved 
\rg{"improved" over what? Actually, we earlier claimed that Juice/Wine did NOT provide accuracy assessments?}
quantification through data-set specificity (\eg Juice Screener, Wine Screener) but are generally specific to a single platform and/or spectrometer frequency.
 
There are a handful of software packages that achieve many of the aforementioned features~\cite{autofit2,ceed2,lcmodel,batman}.  
However, except for LCModel (which can profile only a small set of $\sim 20$ compounds in various tissues),
these software packages still require manual preprocessing and thus are not fully automatable. 
Moreover their use is limited to mixtures with $20-40$ compounds. 

To summarize, \bayesil\ is the only 1D \hone~\NMR\ interpretation system that is
completely automated (both preprocessing and deconvolution) for a wide range of complex mixtures (\ie all mammalian biofluids covered by its current library; see \textit{www.bayesil.ca}), involving $>60$ compounds. Moreover, it is efficient, general, accurate and publicly available.
}
\comment{
There have been a number of software packages 
recently
developed to facilitate spectral profiling and compound identification/quantification by \NMR.  Several were developed specifically for analyzing 2D \NMR\ spectra (\cite{colmar, metabominer}) 
but because of the lengthy time required to collect 2D \NMR\ spectra, most efforts have been focused on analyzing 1D \NMR\ spectra. 
While many 1D analysis packages have been described \cite{ceedthesis,colmar,metabominer,autofit,bquant,metabohunter,batman, newmethod, crockford, nmrglue,ceed}, most have been tested only under very limiting or simplified conditions.  
Some software packages, such as \textsc{NMR-Glue}, facilitate access to different file formats and provide basic preprocessing. In fact \bayesil\ relies on \textsc{NMR-Glue} to read and write all of its~\cite{nmrglue}.
Typically the tests or assessments of these software packages have been restricted to very simple defined mixtures of 5-10 compounds. Some have involved assessments using spike-in experiments with 5-10 compounds added to a biofluid, 
 while others have conducted tests 
or analyses that were restricted to the quietest regions of a biofluid spectrum.
Assessments regarding the performance and accuracy of these systems on real biofluids indicates that they are not particularly accurate (\eg up to $50\%$ false positives for \cite{bquant}). 
Furthermore, many of the packages are not publicly available or appear to be no longer supported \cite{ceed, crockford, autofit, newmethod}.
\refTable{table:software} enumerates the \NMR-analysis software packages and some of their features.
}
\comment{
To date, the largest number of metabolites that has been automatically identified and quantified using published software is 26 compounds (\BATMAN~\cite{batman}). 
However, an analysis of this magnitude required several hours of CPU time to process a single spectrum.  
Furthermore, \BATMAN\ also requires a human expert to perform many of the preliminary steps:
manual transformation, phasing, apodization, baseline correction, water removal and chemical shift referencing.
We compared \BAYESIL\ to \BATMAN\ on simple computer-generated mixtures, involving 5, 10 and 20 compounds selected from \BATMAN's library, as well as a preprocessed serum sample. For the computer generated spectra, both \batman\ and \bayesil\ used the identical library and identified all the compounds. \batman\ achieved $85-87\%$ quantification accuracy for the mixtures with 5, 10, and 20 compounds but took 2-9 hours to run, while in all cases \bayesil\ was $>98\%$ accurate and took less than 3 minutes to run. For the serum spectra, \bayesil\ used a library of 50 compounds, while \batman\ used a subset of 40 compounds that its library has in common with \bayesil. \batman\ took 19 hours to analyze the serum spectrum and achieved and identified all the compounds in its library (resulting in $85\%$ identification accuracy) and $8\%$ quantification accuracy compared to \bayesil\, which took 5 minutes to achieve $98\%$ identification and $87\%$ quantification accuracy. 
}

\appendix[II. \bayesil's loss function]
In the manuscript, we described a simple sum of squared error loss function (\refEq{eq:loss}) to explain the basic ideas behind \bayesil's spectral deconvolution.
In practice \bayesil\ uses a more complicated loss function that also penalizes the derivatives of the difference
between the given spectrum $\spec(\cdot)$ and its reconstruction $\predSpec{\concs, \shifts}{\cdot}$:
{\small
\begin{equation*}
\loss_{\Y}\left(\spec(\cdot), \predSpec{\concs,\shifts}{\cdot}\right) =
\!\!\!\sum_{c \in \{0,1,2,3\}}\gamma_c \int_{\Y}\bigg ( \frac{\partial^c}{\partial \y^c}(\spec(\y)-\predSpec{\concs,\shifts}{\y}) \bigg )^2 \, \mathrm{d}\y
\end{equation*}}

\noindent
where the integral for $c = 0$ corresponds to 
sum of squared errors (\refEq{eq:loss}),
and $c \geq 1$ enforce the smoothness of the difference between $\spec(\cdot)$ and $\recspec(\cdot)$. 
Here, the scalars $\gamma_c$ weight the relative importance of
these terms. 
Similar to the sum of square errors, 
this loss function also decomposes
over the regions $\Y_I$, allowing for the same kind of factor-graph representation.

\appendix[III. Details about the construction of NMR spectral regions]
The construction of the spectral regions is based on the observation that the influence of each peak 
(and hence of each spectral cluster)
is significant over only a relatively small region of the spectrum.
To estimate this ``region of effect'' for each cluster, \bayesil\ first obtains an upper-bound
$\overline{\conc}_{\compound} \geq \conc_\compound$
on the concentration of each compound. This upper-bound is also used in performing approximate
inference (see Appendix~IV).
For each compound $\compound$, this is the minimum of the upper-bounds obtained using each of its clusters $\cluster \in \compound$:
$\overline{\conc}_{\compound} = \min_{\cluster \in \compound} \overline{\conc}_{\cluster}$.  
The upper-bound from each cluster ($\overline{\conc}_{\cluster}$) is obtained by progressively shifting the signature for that cluster under 
the spectrum and finding the maximum value that it can take, assuming all the other compounds and clusters are absent: 
\begin{align*}
\overline{\conc}_{\cluster}\quad =\quad \max_{\shift_\cluster} \min_{\y} \frac{\spec(\y)}{\spec_{\cluster}(\y; 1, \shift_\cluster)}
\end{align*}
where $\spec_\cluster(.,1,\shift_\cluster)$ is the signature of cluster $\cluster$ as defined by the set of all its peaks (see \refEq{eq:signature}) assuming a unit concentration and 
allowing 
$\shift_\cluster$ to vary in a small window $[\underline{\shift}_\cluster, \overline{\shift}_\cluster]$ around the center defined by the library.

Now that we have an upper-bound on concentrations, using an example from \refFigure{fig:blocks} (in the main manuscript) we show why the region of effect for each cluster is bounded.
In this figure, the center ($\shift_{L-Isolucine(1)}$) for the first cluster of {\em L-Isoleucine} can only appear in
the interval [0.9130, 0.9380] \PPM; as its concentration is at most $\overline{\conc}_{L-Isoleucine}\ =\ $95\uM, its contribution to any point outside [.8563, .9954] will be five times less than the estimated {\em noise-level} of this spectrum,
where the noise-level
is estimated as the standard deviation of 
the spectrum $\spec(\cdot)$ over all of the baseline points.
We can therefore identify this L-Isoleucine cluster with the interval [.8563, .9954].
Note this cluster includes 3 peaks.
In general, the range of a cluster spans the set of peaks that it contains.

\appendix[IV. Details of \bayesil's approximate inference]
This appendix provides details of \bayesil's inference procedure,
using 
the factor-graph representation of the problem.

Recall that $\concs_\I$ and $\shifts_I$ denote the set of shift and concentration values for all the clusters and associated compounds that can appear in the region $\Y_I$;
we let 
$\vars_I = [\concs_I , \shifts_I]$
denote the set of variables of both both types.
Since the loss function $\loss(\cdot,\cdot)$ is additive over domain the $\Y$, we can rewrite the optimization of \refEq{eq:optimization} in exponential form as:
\begin{align}  
\vars^* \;     &= \; \arg_{\var}\max \; \prod_I \ff_I(\vars_I) \nonumber\\
\ff_I(\vars_I) &= \expp{-\frac{1}{T}\loss_{\Y_I}(\,\spec(\cdot),\, \recspec(.; \vars_I)\,)}
\label{eq:optsum}
\end{align}
where each factor $\ff_I$ is basically the exponential of the negative loss function
($-\ell_{\Y_I}$) over the region $\Y_I$ and $T$ is the temperature.
Here a factor node $\ff_I$ is connected to all 
of its associated variables $\var_i \in \var_I$ in the factor-graph;
see \refFigure{fig:pgm}.
In the following,
we use $\partial \var_i = \{ \ff_I \; |\; \var_i \in \vars_I\}$ to refer to 
all the factors that are adjacent to variable $\var_i$ in the factor-graph.

Inference in this factor-graph is challenging as its factors 
can each depend on a large number of continuous variables.
This means that the most basic task of (conditional) \textit{sampling} 
from a factor is unfeasible and we cannot use Glauber dynamics (\eg~\cite{glauber}).
This is further complicated by multi-modality of the factors, which  
prevents the use of parametric densities and inference techniques such as Gaussian Belief Propagation~\cite{pgm}
or (primal and dual) decomposition methods that require convex sub-problems \cite{boyd}.
\bayesil\ uses a non-parametric sequential Monte Carlo method that is closely related to sequential importance sampling and particle filters. The following is a step-by-step explanation of this inference procedure.

\bayesil\ models a distribution $\pp(\var_i)$ over individual variables, non-parametrically: 
as a set of ``particles''. These particles, $\var_i\nn{n}$ for $1\leq n \leq N$, collectively serve to approximate the target distribution.
In all the experiments described in the manuscript we use $N=10,000$ such particles, however using a larger number can increase  accuracy at the cost of increased run-time. 

For each $n$, the joint set of particles for all variables, $\vars\nn{n} = [\concs\nn{n},\shifts\nn{n}]$, 
corresponds to a complete spectrum -- \ie $\predSpec{\concs\nn{n},\shifts\nn{n}}{\y}\ =\ 
\sum_{\compound} 
\predSpec{\compound,\conc_\compound\nn{n},\shifts_\compound\nn{n}}{\y}$ -- 
which means we can compute its loss (\refEq{eq:loss}). 
\bayesil\ calculates the loss for each region $\Y_I$,
$\ff_I(\var_I\nn{n})$ (\refEq{eq:optsum}).
Since each variable $\var_i$ appears in many regions
$\factorOf{\var_i}$,
we can ``credit'' each assignment $\var_i\nn{n}$ with the loss over all such regions.
This allows us to compute a ``weight'' 
 for each variable $\var_i$ and for each particle $n$.
\bayesil\ then uses these weighted sets of particles 
to produce a new distribution for the variable $\var_i$,
one that prefers values that have less loss.
\bayesil\ then iterates, using this new distribution,
until convergence.

More specifically, \bayesil\ first assigns each of the variables $\var_i$ to an initial distribution of values $\pp\ttt{0}(\var_i)$
--
\eg $\shift_{L-Isoleucine(1)}$ is drawn uniformly from its chemical shift range [0.9130, 0.9380] \ppm,
and $\conc_{L-Isoleucine(1)}$ is drawn uniformly from its range [0, 95] \uM\
(see Appendix~III for the procedure to estimate the upper bound 
$\overline{\conc}_{L-Isoleucine(1)} = 95$\uM).
It then iterates $t=0,1,2, ...$ over the following four steps:
\\
{\bf Step1:} It draws $N=$10,000 particles from each
$\pp\ttt{t}(\var_i)$ independently, 
producing 10,000 complete assignment to these variables 
$\vars\ttt{t}\nn{n}\ =\ [\concs\ttt{t}\nn{n},\, \shifts\ttt{t}\nn{n}]$ (for $n = 1,2,..., $10,000), from its current product distribution.
%
\\[1ex]
{\bf Step2:} For each joint particle $\vars\ttt{t}\nn{n}$:
For each factor $I$, \bayesil\ computes the loss associated with this region,
$\loss_I(\vars_I\nn{n}\ttt{t})$, and then $\ff_I(\vars_I\nn{n}\ttt{t})$ using 
\refEq{eq:optsum}.
\\[1ex]
{\bf Step3:} Recall each variable $\var_i$ belongs to a set of regions (each corresponding to a factor), $\partial \var_i$.
\bayesil\ then implicitly identifies $\var_i\nn{n}\ttt{t}$ with the loss associated with all corresponding regions. This is achieved by defining a weight
\begin{align}\label{eq:update}
\w(\var_i\nn{n}\ttt{t}) \propto \frac{\prod_{\ff_I \in \partial \var_i} \ff_I(\vars_I\nn{n}\ttt{t})}{\pp\ttt{t}(\var_i\nn{n}\ttt{t})}
\end{align}
where $\pp\ttt{t}(\cdot)$ is the distribution used for sampling
-- \ie importance sampling weight. 
\\[1ex]
{\bf Step4:} Finally, \bayesil\ produces a ``new'' distribution for each variable $\var_i$,
using kernel density estimation (KDE) over a weighted set of the $N$ 
particles $\vars\nn{n}\ttt{t}\; 1 \leq n \leq N$ to represent the marginal 
distribution $\pp\ttt{t+1}(\cdot)$ over each variable $\var_i \in \vars$:
\begin{align*}
\pp\ttt{t+1}(\var_i)\ \propto\ \sum_{n=1}^{N} \w(\var_i\nn{n}\ttt{t}) \
\kk\left(  \frac{\var_i - \var_i\nn{n}\ttt{t}}{h} \right)
\end{align*}
where $\kk(\cdot)$ is a kernel function (\eg a Gaussian) 
and the kernel bandwidth $h$ is estimated from the data (\eg \cite{kde}).\\[1ex]
\bayesil\ then checks for convergence; if convergence occurs, it returns the mode of individual distribution as its approximation to 
the MAP assignment.
If not, it returns to Step1.

Note the temperature parameter $T$ (used in \refEq{eq:optsum} which appears in \refEq{eq:update}) is gradually reduced from a large value towards zero per iteration.
\refFigure{fig:kde} is basically showing the evolution of KDEs over the chemical shift variables ($\pp\ttt{t}(\shift_\cluster) \; \hbox{for}\ t \in \{1,\ldots,6\}$) 
over $6$ iterations of spectral deconvolution.

In practice, \bayesil\ ignores the importance sampling weights. 
This biases $\pp\ttt{t+1}(\var_i)$ towards $\pp\ttt{t}(\var_i)$, but significantly reduces the variance and the computation time.
Since we are interested
in the mode (rather than the marginals) of $\pp(\var_i)$, this trade-off is favourable,
as 
the bias of the previous estimate
is mostly towards more probable assignments.

\appendix[V. List of NMR-detectable compounds in Serum and CSF]
Our serum library includes the following metabolites plus DSS:\\
1-Methylhistidine, 
2-Hydroxybutyric acid, 
Acetic acid, 
Betaine, 
Acetoacetic acid,
L-Carnitine,
Creatine,
Citric acid,
Choline,
Ethanol,
D-Glucose,
Glycine,
Glycerol,
Formic acid,
L-Glutamic acid,
Hypoxanthine,
L-Tyrosine,
L-Phenylalanine,
L-Alanine,
L-Proline,
L-Threonine,
L-Asparagine,
L-Isoleucine,
L-Histidine,
L-Lysine,
L-Serine,
L-Lactic acid,
L-Aspartic acid,
L-Cystine,
Ornithine,
Pyruvic acid,
Succinic acid,
Urea,
3-Hydroxybutyric acid,
L-Arginine,
Creatinine,
L-Cysteine,
L-Glutamine,
L-Leucine,
Malonic acid,
L-Methionine,
Isopropyl alcohol,
L-Valine,
L-Tryptophan,
Acetone,
Isobutyric acid,
Methanol,
Propylene glycol,
Dimethyl sulfone.
\\[2ex]
\noindent
Our CSF library includes the following compounds plus DSS:\\
2-Hydroxybutyrate,
2-Oxoisovalerate,
3-Hydroxyisobutyrate,
Acetate,
Ascorbic acid,
Acetoacetate,
Creatine,
Dimethylamine,
Citrate,
Choline,
Glucose,
Glycerol,
Formate,
Glutamate,
Tyrosine,
Phenylalanine,
Alanine,
Threonine,
Mannose,
Isoleucine,
Histidine,
Lysine,
Serine,
Lactate,
2-Oxoglutarate,
myo-Inositol,
Oxalacetate,
Pyruvate,
Succinate,
Pyroglutamate,
Xanthine,
Urea,
3-Hydroxybutyrate,
2-Hydroxyisovalerate,
Creatinine,
Glutamine,
Fructose,
Leucine,
Methionine,
3-Hydroxyisovalerate,
Isopropanol,
Valine,
Tryptophan,
Acetone,
Methanol,
Propylene glycol,
1,5-Anhydrosorbitol,
Dimethylsulfone.

\comment{
\begin{figure*}
\centering
\includegraphics[width=.9\linewidth]{fit_serumcsf3.pdf}
\caption{Part of the fit produced by \bayesil\ for sysnthetic serum+CSF sample. Clusters of the same compound are in the same color.}
\label{fig:fit_serumcsf}
\end{figure*}

\begin{figure*}
\centering
\hbox{
\includegraphics[width=.45\linewidth]{synthetic_biological_2.pdf}
\includegraphics[width=.45\linewidth]{simulated_2.pdf}
}
\caption{(left) average identification and quantification accuracy of \bayesil\ on synthetic 
and biological, serum, CSF 
(and the synthetic mixture of CSF and serum) samples.
(right) average identification and quantification accuracy of \bayesil\ versus
human expert in profiling computer-generated serum and CSF spectra where the ground-truth
is known.}
\label{fig:error}
\end{figure*}

\comment{
\begin{figure*}
\centering
\hbox{

}
\caption{
(right) The cumulative histogram of relative error in profiling computer-generated serum samples by both \bayesil\ and human expert. For example the plot shows that \bayesil\ 90\% of times has more than 90\% quantification accuracy.}
\label{fig:threshold}
\end{figure*}
}

\begin{figure*}
\centering
\hbox{
    \includegraphics[width=.4\linewidth]{threshold_median.pdf}
  \includegraphics[width=.4\linewidth]{scatter_serum_10.pdf}
}
\caption{
(left) The effect of detection threshold on identification and quantification accuracy in biological serum samples. The values below the threshold
are assumed absent also in the ground truth. The quantification accuracy reflects the relative accuracy on correctly identified compounds. 
(right) The scatter-plot comparing the quantification of 10 biological serum samples in \bayesil\ vs. experts. Here each sample is quantified by two experts from the same lab and the size of each disk is proportional to the disagreement between expert quantifications.
}
\label{fig:extra_serum}
\end{figure*}

\appendix[V. more details on experimental results]
This appendix gives more details on \bayesil's assesment.
\refFigure{fig:error} visualizes the results of \refTable{table:accuracy}.
\refFigure{fig:threshold} shows the effect of changing the detection threshold on both
identification and quantification accuracy for biological serum samples.

\begin{figure*} 
\includegraphics[width=1.0\linewidth]{Pipeline.png}
\caption{SHOULD WE INCLUDE THIS? If so.. .where?  
DRAFT of the pipeline.  just a sketch \sr{maybe someone wants to produce this for  the web-page?}
}
\end{figure*}
}










\end{document}